%% file: root.tex
\documentclass[letterpaper, 10 pt, conference]{ieeeconf}  
\usepackage{graphicx}
\usepackage{amsmath}
\usepackage{amssymb}
\usepackage{cite}
\usepackage[hidelinks]{hyperref}
\usepackage{tabularx}
\usepackage{array}

\IEEEoverridecommandlockouts                              

\title{\LARGE \bf
TapTacTouch: Shape Completion from Sparse Tactile Contacts with Geometric Priors
}
\author{
  Yoobin Shin$^{1}$, Jaeyoung Lee$^{1}$, Jieun Kang$^{1}$, Hyun Song$^{1}$, Woojin Kim$^{2}$, and Andrew Jaeyong Choi*$^{1}$ \\
  \small $^{1}$School of Computing, iRASC Lab., Gachon University \\
  \small $^{2}$Department of Electronic Engineering, Namseoul University \\
  \small \texttt{\{toy7552, ljy0111, kje147459, songhyun, andrewjchoi\}@gachon.ac.kr} \\
  \small \texttt{crayon922@nsu.ac.kr}
}

\begin{document}

\maketitle
\thispagestyle{empty}
\pagestyle{empty}

\begin{abstract}

3D object reconstruction is essential for robotic perception and manipulation, yet conventional approaches predominantly rely on visual observations that can be limited by occlusion or unavailable in visually constrained environments. In this work, we propose a vision-free, training-free, and category-agnostic framework for reconstructing 3D object meshes from sparse tactile contacts without relying on pretrained shape models. A 6-DoF robotic arm equipped with a 20-DoF multi-finger robotic hand collects tactile observations using 3 × 3 pressure sensor arrays mounted on the finger pads. The acquired tactile measurements are transformed into 3D contact constraints through robot kinematics and integrated into a multi-scale Gaussian Process Implicit Surface (GPIS). To address sparse and incomplete surface observations, the proposed method incorporates contact-confidence weighting, weak pseudo-surface constraints, and an observation-driven geometric prior to preserve local surface geometry and provide structural guidance in unobserved regions. Real-world experiments on six objects achieve a Chamfer Distance (CD) of 1.610, an Earth Mover’s Distance (EMD) of 7.738, and an F-Score of 95.00, with reconstruction completed in an average of 49.99 s per object. These results demonstrate that sparse tactile sensing can provide sufficient geometric information for continuous 3D object reconstruction without visual or semantic information.

\end{abstract}

\section{INTRODUCTION}

\input{1_introduction}

\section{Related Work}
\input{2_related_work}

\section{METHOD}
\input{4_method_overview}
\input{5_method_scanNpolicy}
\input{6_method_reconstruction}

\section{Experiments}

\input{7_experiment_HW_setup}
\input{8_expriment_object}
\input{9_experiment_reconstruction}

\section{CONCLUSIONS}

\input{11_conclusion}





\section*{APPENDIX}

\input{12_Appendix}


\begin{figure}[h]
    \centering
    \includegraphics[width=\columnwidth]{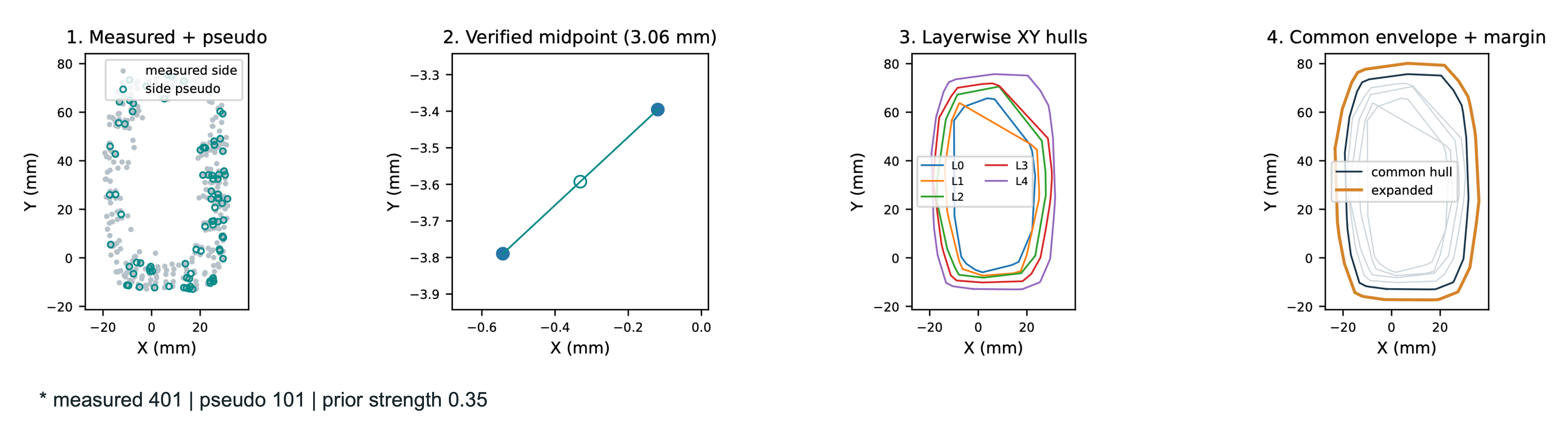}
    \caption{PCG and CSP construction for the object 3 object:
contact-pair interpolation, layer-wise hulls, and the expanded
common cross-section.}
    \label{fig:appendix_results}

\end{figure}

\renewcommand{\thefigure}{A\arabic{figure}}
\setcounter{figure}{0}


\bibliographystyle{IEEEtran}
\bibliography{references}

\end{document}

%% file: 1_introduction.tex
3D object reconstruction is an important component of robotic perception and manipulation, enabling robots to understand object geometry for downstream tasks such as grasping, interaction, and shape-aware manipulation \cite{c1,c2}. Many existing 3D reconstruction approaches rely primarily on visual observations, including RGB and depth images, to recover object geometry \cite{c1}. Although vision provides rich global information, its effectiveness can be limited in real-world manipulation scenarios. In particular, parts of an object may be occluded by the robotic hand, surrounding objects, or the object itself, making it difficult to obtain complete geometric information from visual observations alone \cite{c3,c4}. This limitation becomes more pronounced in cluttered environments, where multiple objects may further restrict camera visibility \cite{c3}.

Tactile sensing provides an alternative means of perceiving object geometry through direct physical interaction. Unlike vision, tactile sensing does not rely on line-of-sight visibility and can acquire local contact information from regions that are difficult to observe from external viewpoints \cite{c1,c2,c5}. These characteristics make tactile sensing particularly suitable for robotic manipulation scenarios in which visual information is limited or unavailable \cite{c4,c6}.

\begin{figure}[!t]
    \centering
    \includegraphics[width=\columnwidth]{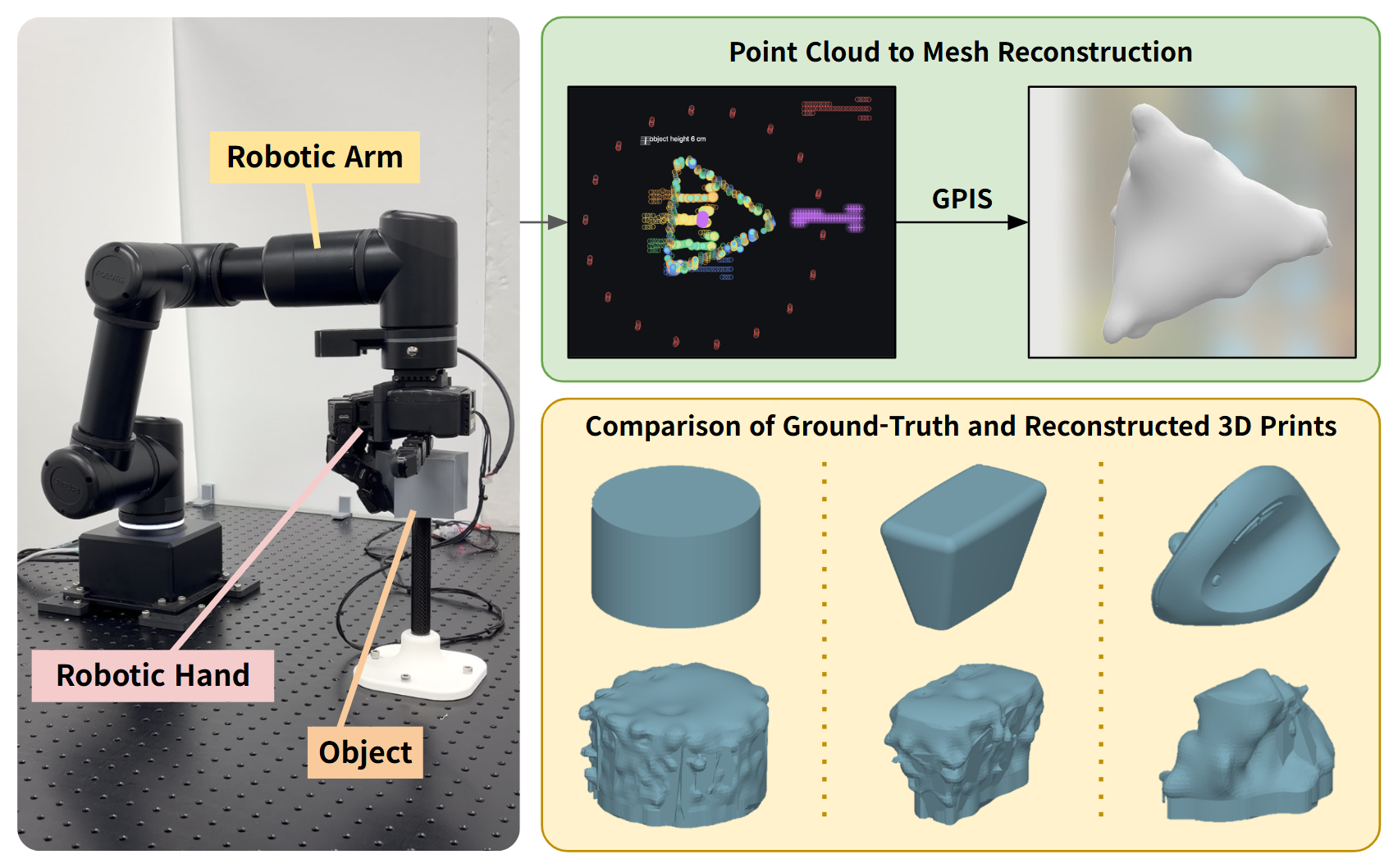}
    \caption{Overview of the proposed tactile-only 3D reconstruction framework.
Tactile contact points are aggregated into a 3D point cloud and reconstructed
into a 3D mesh using Gaussian Process Implicit Surfaces (GPIS).}
    \label{fig:flow}
\end{figure}

In this work, we investigate whether a robot can reconstruct the 3D geometry of an object using tactile data alone, without relying on external cameras or explicit object class labels. We propose a vision-free 3D mesh reconstruction framework using a 6-DoF robotic arm equipped with a 20-DoF right robotic hand. During interaction with an object, tactile data are collected from sensors mounted on the finger pads of the robotic hand and used to infer the object geometry. We employ Gaussian Process Implicit Surfaces (GPIS) \cite{c7,c6} to reconstruct a continuous object surface from the acquired tactile contacts and extract its 3D mesh. Since GPIS does not require data-driven model training, the proposed framework can reconstruct previously unseen objects directly from tactile measurements without a separate training stage or explicit object category information. Fig.~\ref{fig:flow} illustrates the overall pipeline from tactile data acquisition and point-cloud construction to GPIS-based 3D mesh reconstruction.

We evaluate the proposed approach on a physical robotic system using tactile data collected through direct interactions with real objects. The proposed framework reconstructs 3D object meshes from sparse tactile contacts without requiring visual input, prior knowledge of the object class, or model training. Once the tactile contacts are acquired, the GPIS-based reconstruction generates a 3D mesh in approximately 20 s, demonstrating that object geometry can be recovered efficiently without a separate training stage. Our real-world experiments demonstrate the feasibility of training-free, tactile-only 3D reconstruction and highlight the potential of tactile sensing for object reconstruction in scenarios where visual sensing is limited or unavailable.

The main contributions of this work are as follows:
\begin{itemize}
\item We propose a vision-free 3D object reconstruction framework that reconstructs object meshes using tactile data alone.
\item We develop a tactile data acquisition pipeline using a 6-DoF robotic arm and a 20-DoF robotic hand equipped with finger-pad tactile sensors.
\item We validate the proposed framework on a physical robotic platform using real objects, demonstrating the feasibility of tactile-only 3D mesh reconstruction in real-world settings.
\end{itemize}

%% file: 2_related_work.tex
\label{sec:Related Work}

3D object reconstruction has primarily been developed based on visual observations such as RGB or depth images. However, vision alone has limitations in capturing occluded regions and local surface geometry that is difficult to observe. Tactile sensing has therefore been employed as a complementary modality that acquires local surface information through direct physical contact. Accordingly, vision–touch reconstruction methods have been developed to combine global visual information with local tactile measurements, demonstrating improved reconstruction performance over either modality alone \cite{c2,c17}. Nevertheless, these methods still depend on visual observations, limiting their applicability when visual information is degraded or unavailable. During robotic manipulation, regions of interest can also be occluded by the hand or the object itself. These limitations have motivated tactile-only approaches that acquire object shape information solely through physical contact without relying on vision \cite{yi2016active,c19}.

\subsection{Hardware}

The type of tactile sensor determines the form and amount of information obtained from each contact. Recent systems widely employ vision-based tactile sensors such as GelSight, DIGIT, and TacTip, which capture dense local contact information by imaging deformations of the sensing surface \cite{c2,c17,c5}. Compact sensors have also been developed for integration with multi-fingered robotic hands; for example, DIGIT was specifically designed with a compact form factor that enables tactile perception and manipulation with multiple fingers \cite{c2}. However, these sensors produce high-dimensional tactile images, requiring additional processing or learned inference to extract contact locations or local geometric information for 3D reconstruction \cite{c2,c17,c5}.

For tactile-only perception, obtaining a sufficient number of contacts during exploration is important for observing a broad region of an object. Earlier Gaussian Process (GP)-based tactile exploration methods primarily acquired contact observations sequentially \cite{c18,c19}, whereas multi-finger robotic hands can obtain multiple contact observations through several fingers during a single interaction. TactoFind \cite{c23} demonstrates this capability using a three-finger hand, recognizing objects without vision by representing tactile observations using binary contact signals and fingertip positions rather than the full tactile images. This suggests that contact locations accumulated through multi-finger exploration can themselves serve as useful geometric observations without requiring dense local geometry to be inferred from tactile images.

In this work, we use 3$\times$3 pressure arrays mounted on the finger pads of a multi-finger robotic hand. Unlike vision-based tactile sensors that require geometric information to be extracted from tactile images, the pressure arrays directly provide numerical contact signals at sensing elements with known locations. We therefore transform the locations of active sensing elements into 3D contact points through the forward kinematics (FK) of the arm–hand chain and use them as geometric observations for 3D shape reconstruction. This allows contact information acquired from multiple fingers to be directly used for reconstruction without additional tactile-image processing or learned geometry inference.

\subsection{Reconstruction}

Recent tactile-based 3D reconstruction methods have employed high-resolution information from vision-based tactile sensors together with learned reconstruction models. TouchSDF extracts CNN-based features from tactile images and combines them with DeepSDF to reconstruct continuous 3D shapes \cite{c5}, while Touch2Shape employs a touch-conditioned diffusion model to generate object shapes from sparse tactile observations \cite{c21}. TouchAnything further leverages a pretrained visual diffusion model together with sparse tactile contacts and class-level descriptions for shape estimation \cite{c22}. Although these approaches can infer object shape from limited tactile observations using learned models, they rely on pretrained models or training data.

In contrast, tactile-only reconstruction methods have investigated estimating object surfaces directly from physical contacts without relying on visual observations or learned shape models. Early approaches employed Gaussian processes and GPIS to represent sparse contact observations as continuous implicit surfaces and combined them with active tactile exploration to progressively reconstruct object surfaces \cite{c18,c19,c24,c25}. However, because physical contact is inherently limited to accessible regions, insufficiently observed regions remain uncertain, making accurate reconstruction of the complete object shape from sparse contacts challenging.

To address this problem, our approach directly incorporates sparse 3D contact observations acquired through multi-finger pressure sensing into a training-free GPIS framework. We further complement shape estimation in unobserved regions with a geometric prior, enabling continuous 3D mesh reconstruction without model training or explicit object class information.

%% file: 4_method_overview.tex
\label{sec:METHOD}

\subsection{Overview}

\begin{figure*}[t]
    \centering
    \includegraphics[width=\textwidth]{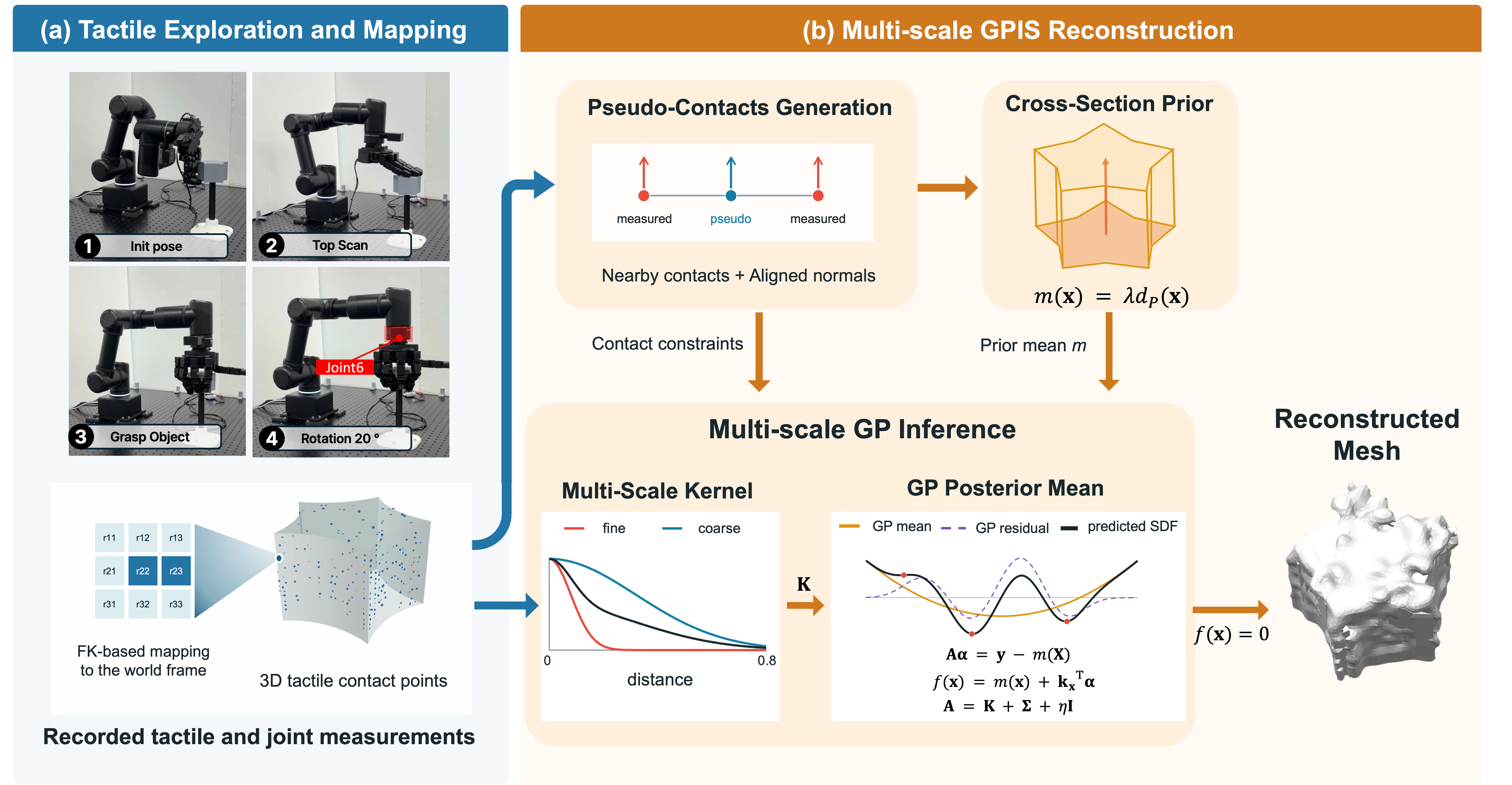}
    \caption{Overview of the proposed tactile reconstruction pipeline.
    (a) An robotic arm equipped with an robotic hand collects
    tactile contacts and kinematic information, which are
    mapped into world-frame contacts and FK-derived normals.
    (b) Pseudo-Contact Generation (PCG) augments the observations,
    and the Cross-Section Prior (CSP) defines the GP mean.
    Multi-scale Gaussian regression estimates a residual field
    that adjusts this mean to the tactile constraints.
    Marching Cubes extracts the reconstructed mesh from the
    zero-level set of the resulting implicit field.}
    \label{fig:overview}
\end{figure*}

We reconstruct 3D object geometry from tactile observations and kinematic information acquired through robotic contact exploration. An robotic arm equipped with an robotic and sequentially performs lower and upper boundary detection, top-surface scanning, and side-surface scanning. The observed contact heights and the distribution of top-surface contacts determine the vertical range and horizontal reference position for subsequent side-surface scanning.
The collected contacts and FK-derived normals are augmented through Pseudo-Contact Generation (PCG) and converted into signed surface constraints. A Cross-Section Prior (CSP) defines the GP mean, while multi-scale Gaussian regression estimates a residual field that adjusts this mean to the tactile constraints. The final mesh is extracted from the zero-level set using Marching Cubes~\cite{c13}.

%% file: 5_method_scanNpolicy.tex
\label{sec:METHOD}
\subsection{Robotic Arm \& Hand Motion}

An OMY robotic arm equipped with an HX5-D20-MRT hand collects contact observations from the top and side surfaces of an object. The arm controls the hand position and approach direction, while the hand adopts configurations that promote sensing-surface contact. Only the four non-thumb fingers are used for contact exploration, as the shorter thumb has limited reach under the selected scanning configurations. The Tool Center Point (TCP) is the end-effector motion-control reference, distinct from tactile contact points. The object is assumed to remain stationary and accessible along the scanning trajectories. A rule-based policy sequentially performs lower and upper boundary detection, top-surface grid scanning, and layered side-surface scanning.

For lower boundary detection, the four fingers flex their distal joints, and the TCP rises in $10~\mathrm{mm}$ increments at a fixed horizontal position. For upper boundary detection, it descends from a non-contact position at the same interval. Tactile signals are measured at each position, and the median $z$-coordinate of active taxels at the first valid contact defines the corresponding boundary height. The height difference determines the number of side-scanning layers. Exploration terminates if no contact is detected within the prescribed interval.

After top contact is detected, the hand adopts the top-scanning configuration, and the TCP rises $25~\mathrm{mm}$ and moves $50~\mathrm{mm}$ along the robot-base $+x$-axis to establish the grid origin. It then traverses a $7 \times 4$ grid with $20~\mathrm{mm}$ spacing along $-x$ and $10~\mathrm{mm}$ along $+y$. At each location, the TCP descends $30~\mathrm{mm}$ from the grid-origin height, records tactile signals for $0.50~\mathrm{s}$, and returns to the same height. This fixed-displacement approach promotes sensing-surface contact to acquire top-surface contact points.

After completing the top scan, the horizontal reference position for side-surface scanning is computed as the arithmetic mean of the valid contact coordinates. For $N>0$ contact points $(x_i,y_i)$ expressed in the robot base frame,

\begin{equation}
\bar{x} = \frac{1}{N}\sum_{i=1}^{N} x_i,
\qquad
\bar{y} = \frac{1}{N}\sum_{i=1}^{N} y_i.
\label{eq:side_scan_reference}
\end{equation}

This reference is used to update the initial side-scanning TCP position and layer configurations. Layers are traversed from top to bottom at $10~\mathrm{mm}$ intervals, with additional upper-layer configurations generated by inverse kinematics (IK) when required. The target objects are up to $60~\mathrm{mm}$ in height and $80$--$100~\mathrm{mm}$ in width. For this size range, 18 Joint6 settings from $-170^{\circ}$ to $170^{\circ}$ in $20^{\circ}$ increments are selected to balance contact opportunities across different finger configurations against scan duration. The sweep direction is reversed between adjacent layers to reduce return motions to the starting angle. At each wrist setting, the hand sequentially adopts initial-contact and predefined flexed configurations.

To supplement finger flexion, the arm applies a horizontal TCP displacement in a predefined direction associated with each wrist setting. For wrist-setting index $k \in \{0,\ldots,17\}$, the target position is defined as

\begin{equation}
\begin{aligned}
\phi_k &= \frac{2\pi k}{18}, \\
\mathbf{p}_k^{\star}
&=
\mathbf{p}_k
+
r
\begin{bmatrix}
\cos\phi_k \\
\sin\phi_k \\
0
\end{bmatrix}.
\qquad
\end{aligned}
\label{eq:tcp_horizontal_displacement}
\end{equation}

where $\mathbf{p}_k$ denotes the pre-approach TCP position in the robot base frame, and $\phi_k$ denotes the horizontal displacement angle, distinct from the Joint6 joint angle. The target specifies a $7~\mathrm{mm}$ horizontal displacement without changing the target height. A Jacobian-based position IK solver computes arm joint targets with Joint6 held fixed, and the targets are executed under position control. Tactile measurements are acquired at rest before the TCP returns to its pre-approach position and the hand releases contact. Each measurement is recorded with the corresponding joint states.

%% file: 6_method_reconstruction.tex
\label{sec:METHOD}

\subsection{Multi-scale GPIS Reconstruction}

We reconstruct a class-free object mesh from sparse tactile contacts using a
training-free multi-scale GPIS~\cite{c7}. As shown in
Fig.~\ref{fig:overview}, tactile contacts and their FK-derived outward normals
are converted into signed surface constraints. A multi-scale Gaussian kernel
interpolates these constraints into a continuous implicit field, and its zero-level
set is extracted as the initial mesh.

Before constructing the GP constraints, we normalize the contact
coordinates using the bounding-box center $\mathbf c$ and scale $s$:
\begin{equation}
\begin{aligned}
\mathbf c &= \frac{\mathbf p_{\min}+\mathbf p_{\max}}{2},\\
L &= \max_k(p_{\max,k}-p_{\min,k}),\\
s &= \max(L/1.6,10^{-6}),\\
\widetilde{\mathbf p}_i &= (\mathbf p_i-\mathbf c)/s.
\end{aligned}
\end{equation}
Hereafter, we omit the tilde and express all positions in
normalized coordinates. All length parameters, including
$\varepsilon$, $\ell_{\mathrm f}$, $\ell_{\mathrm c}$, and the
shared margin $\rho$, are specified in normalized coordinates;
their physical lengths are obtained by multiplying them by $s$.

A Signed Distance Function (SDF) is positive outside an object, negative inside,
and zero on its surface~\cite{c8}. We assign a zero target to every measured or generated pseudo-contact.
For each contact position $\mathbf{p}_i$ and outward normal $\mathbf{n}_i$, we construct
\begin{equation}
\begin{aligned}
\mathbf{p}_i^{\pm} &= \mathbf{p}_i \pm \varepsilon\mathbf{n}_i, \\
f(\mathbf{p}_i) &= 0, \qquad
f(\mathbf{p}_i^{+})=+\varepsilon, \qquad
f(\mathbf{p}_i^{-})=-\varepsilon,
\end{aligned}
\label{eq:tactile_constraints}
\end{equation}

where $\varepsilon=0.012$. These constraints encourage the reconstructed zero-level set
to remain close to the measured contacts while encoding the
local inside--outside orientation. We also assign positive SDF values to samples on the reconstruction
boundary and a negative value to the mean contact position. Under the assumption
of a single solid object containing the contacts, these auxiliary constraints
provide weak outside and inside supervision and encourage surface
closure.

Following the Gaussian covariance formulation commonly used in Gaussian
processes~\cite{c9}, we combine fine- and coarse-scale RBF
kernels as
\begin{equation}
\begin{aligned}
k(\mathbf{x},\mathbf{x}')
&= w\exp\!\left(-\frac{\lVert\mathbf{x}-\mathbf{x}'\rVert_2^2}
{2\ell_{\mathrm f}^2}\right) \\
&\quad +(1-w)\exp\!\left(-\frac{\lVert\mathbf{x}-\mathbf{x}'\rVert_2^2}
{2\ell_{\mathrm c}^2}\right),
\end{aligned}
\label{eq:multiscale_kernel}
\end{equation}
where $\ell_{\mathrm f}=0.08$, $\ell_{\mathrm c}=0.32$, and $w=0.5$.
The fine scale represents local variations near contacts, whereas the coarse
scale supplies long-range correlation between spatially separated
observations~\cite{c9}. The following subsections describe pseudo-contact generation
and the construction of the cross-section prior, which are
combined in the subsequent GP inference stage.

\subsection{Pseudo-Contact Generation (PCG)}

Before generating pseudo-contacts, we assess the reliability
of measured side contacts using local geometric consistency.
Within each acquisition layer, we define a reference center
as the coordinate-wise median of the side-contact $xy$ positions
and divide the surrounding plane into 24 equal angular sectors.
The neighborhood of each contact comprises its own sector
and the two immediately adjacent sectors in the same layer.

For contact $i$, let $r_i$ be its radial distance from the
layer reference center, and let $\bar r_i$ and
$\mathrm{MAD}_i$ denote the median and median absolute
deviation of the neighborhood radii. We define
\begin{equation}
\begin{aligned}
t_i &= \max\!\left(
1.5 \times 1.4826\,\mathrm{MAD}_i,\;0.006/s
\right),\\
R_i &= \operatorname{clip}_{[0,1]}
\left(1-\frac{|r_i-\bar r_i|}{t_i}\right),\\
O_i &= \operatorname{clip}_{[0,1]}
\left(\frac{a_i-0.35}{0.65}\right),\\
U_i &= \operatorname{clip}_{[0,1]}
\left(\frac{M_i-1}{2}\right),\\
q_i &= \frac{0.45R_i+0.25O_i+0.20U_i}{0.90}.
\end{aligned}
\label{eq:contact_reliability}
\end{equation}
Here, $a_i$ is the dot product between the unit
$xy$-projected normal and the unit outward radial direction,
and $M_i$ is the neighborhood contact count, including
contact $i$. We set $a_i=0$ if either direction is undefined.
The operator $\operatorname{clip}_{[0,1]}$ clamps its argument
to $[0,1]$. The tolerance floor $0.006/s$ corresponds to
6 mm in physical coordinates, with $s$ expressed in meters
per normalized unit.

Contacts are classified as reliable for $q_i\geq0.6$,
uncertain for $0.3\leq q_i<0.6$, and unreliable otherwise.
Only reliable measured side contacts are used to generate
side pseudo-contacts and construct the side contours of
the CSP. Uncertain and unreliable measured side contacts
remain in the GP observation set, with their surface and
normal-offset noise standard deviations multiplied by
3 and 7, respectively.

We augment the measured contacts with pseudo-contacts
interpolated between nearby observations with consistent
normal directions. Pairs are formed within the top-contact
set and, separately, within the reliable side contacts of
each acquisition layer, using the reliability criterion
defined above. For a pair $(i,j)$, we require
$\lVert\mathbf p_i-\mathbf p_j\rVert_2\leq d_{\mathrm{NN}}$
and $\mathbf n_i^{\mathsf T}\mathbf n_j\geq0.93$,
where $d_{\mathrm{NN}}$ is the median nearest-neighbor
distance within the corresponding contact group.
The pseudo-contact position and normal are
\begin{equation}
\begin{aligned}
\mathbf p_{ij}^{\mathrm p}
&=\frac{\mathbf p_i+\mathbf p_j}{2},\\
\mathbf n_{ij}^{\mathrm p}
&=\frac{\mathbf n_i+\mathbf n_j}
{\lVert\mathbf n_i+\mathbf n_j\rVert_2}.
\end{aligned}
\label{eq:pseudo_contact_generation}
\end{equation}

Eligible pairs are ordered by increasing separation,
and the number of pseudo-contacts is capped at
$\operatorname{round}(0.5N_{\mathrm r})$, where
$N_{\mathrm r}$ is the number of measured contacts.
We assign a surface-zero target to each pseudo-contact,
with observation-noise standard deviations of $0.025$
for top and $0.035$ for side pseudo-contacts in the
normalized implicit field.
Generated points are not reused recursively as measured
contacts. This module reinforces local continuity between
nearby contacts; it does not by itself propagate observations
across broad unsupported regions.

\subsection{Cross-Section Prior (CSP)}

Sparse tactile contacts leave distant regions weakly constrained. We therefore
construct a CSP directly from the current tactile contacts.
We assess whether the observed layers exhibit similar radial profiles and
cross-sectional sizes. When this consistency supports a shared cross-section,
the valid layer contours are enclosed by a common polygonal envelope, which
defines a geometric GP mean over the observed vertical extent with margins.
The polygon is estimated from the input contacts rather than selected from a
predefined collection of shapes.

Reliable measured side contacts and generated side pseudo-contacts provide
the layerwise $xy$ contours used to construct the CSP.
Non-side measurements and top pseudo-contacts contribute to its vertical
extent. In the reliability-filtered configuration, the prior requires at least
20 reliable measured side contacts distributed across two or more acquisition
layers. This support requirement is checked separately from the geometric
consistency score.
For each acquisition layer $\ell$, let $\mathcal S_\ell$ index
the reliable measured side contacts and generated side
pseudo-contacts, jointly denoted by $\mathbf p_i$.
For layers containing at least three non-collinear side points,
we construct a layerwise convex hull and combine the valid
hulls into a common polygonal envelope~\cite{c10}:
\begin{equation}
\begin{aligned}
H_\ell &= \operatorname{ConvHull}
\left(\{\mathbf{p}_{i,xy}:i\in\mathcal{S}_\ell\}\right),\\
P_0 &= \operatorname{ConvHull}
\left(\bigcup_{\ell\in\mathcal{L}_{\mathrm{valid}}}H_\ell\right).
\end{aligned}
\label{eq:layerwise_polygon}
\end{equation}
The common polygon is formed from the vertices of the valid layer hulls. If no valid layer hull
exists, we use the convex hull of all available side points in the prior input.
If that hull is also degenerate, we use a 16-vertex circular approximation.
To provide an outward margin around the contact-derived contour, we expand
its vertices radially by $\rho=0.08$ in normalized coordinates from the center $\mathbf c_{xy}$ of
the side-contact bounding box:
\begin{equation}
\mathbf{v}'_j=\mathbf{v}_j+\rho
\frac{\mathbf{v}_j-\mathbf{c}_{xy}}
{\max(\lVert\mathbf{v}_j-\mathbf{c}_{xy}\rVert_2,10^{-12})}.
\label{eq:radial_polygon_expansion}
\end{equation}
The expanded vertices define the polygon $P$. This radial expansion is not a
constant-distance polygon buffer because the displacement direction is measured
from $\mathbf{c}_{xy}$ rather than perpendicular to each edge.

Using all points admitted to the mean-function input, the vertical center and
half-height are
\begin{equation}
z_c=\frac{z_{\min}+z_{\max}}{2}, \qquad
h=\frac{z_{\max}-z_{\min}}{2}+\rho,
\label{eq:cylinder_height}
\end{equation}
where $\rho=0.08$. Let $d_{xy}(\mathbf{x})$ be the signed distance from the
$xy$ projection of $\mathbf{x}$ to $P$, with negative values inside and positive
values outside. Its magnitude is obtained from the minimum point-to-segment
distance, and its sign is determined by a ray-crossing test~\cite{c11}.
The vertical signed distance is $d_z(\mathbf{x})=|z-z_c|-h$.
The finite polygonal-prism distance is
\begin{equation}
\begin{aligned}
d_P(\mathbf{x})
&=\sqrt{\max(d_{xy},0)^2+\max(d_z,0)^2}\\
&\quad+\min\!\left(\max(d_{xy},d_z),0\right).
\end{aligned}
\label{eq:finite_polygon_prism_sdf}
\end{equation}

We compute a geometric-consistency score $s_P\in[0,1]$ for the
contact-derived polygonal-prism prior using inter-layer profile
and size consistency, aspect ratios, and contact-count support.
The GP mean is defined as
\begin{equation}
\begin{aligned}
m(\mathbf{x}) &= \lambda(s_P)d_P(\mathbf{x}),\\
\lambda(s_P) &=
\begin{cases}
0, & s_P < 0.6,\\
\min(0.5s_P, 0.35), & s_P \geq 0.6.
\end{cases}
\end{aligned}
\label{eq:adaptive_mean_weight}
\end{equation}
If the reliable-side support condition is not met, we set
$m(\mathbf{x})=0$ regardless of $s_P$.
Otherwise, the score determines whether the prior is activated
and scales its mean amplitude, which is capped at $0.35$.

\subsection{Gaussian Process Inference and Mesh Extraction}

Let $\mathbf{X}=[\mathbf{X}_1,\ldots,\mathbf{X}_N]^{\mathsf T}$ contain all
constraint positions and $\mathbf{y}=[y_1,\ldots,y_N]^{\mathsf T}$ their target
SDF values. GPIS represents the implicit surface using the posterior mean of a
GP conditioned on these observations
~\cite{c9,c7}. We
use the multi-scale kernel in Eq.~\eqref{eq:multiscale_kernel} to determine how
strongly each measured or pseudo constraint influences a spatial query. Its
short length scale preserves variations near a contact, while its long length
scale propagates information across weakly observed regions. Both length scales
and their mixing weight remain fixed during reconstruction
~\cite{c9}.

Let $\sigma_i$ denote the observation-noise standard deviation assigned to the
$i$-th constraint. We define the kernel matrix, query kernel vector, and
noise-regularized GP system matrix as
\begin{equation}
\begin{aligned}
K_{ij}&=k(\mathbf{X}_i,\mathbf{X}_j), \\
(\mathbf{k}_{\mathbf{x}})_j&=k(\mathbf{x},\mathbf{X}_j), \\
\mathbf{A}&=\mathbf{K}+
\operatorname{diag}(\sigma_1^2,\ldots,\sigma_N^2)+\eta\mathbf{I},
\end{aligned}
\label{eq:gpis_system}
\end{equation}
where $\eta=10^{-6}$ is numerical jitter. A smaller $\sigma_i$ gives a reliable
constraint greater influence, whereas a larger $\sigma_i$ reduces the influence of
uncertain constraints. The jitter improves numerical stability and
helps keep $\mathbf{A}$ positive definite~\cite{c9}. With
$m(\mathbf{X})=[m(\mathbf{X}_1),\ldots,m(\mathbf{X}_N)]^{\mathsf T}$, the
coefficients and predicted SDF are
\begin{equation}
\begin{aligned}
\mathbf{A}\boldsymbol{\alpha}&=\mathbf{y}-m(\mathbf{X}),\\
f(\mathbf{x})&=m(\mathbf{x})+
\mathbf{k}_{\mathbf{x}}^{\mathsf T}\boldsymbol{\alpha}.
\end{aligned}
\label{eq:gpis_nonzero_mean}
\end{equation}
The right side $\mathbf{y}-m(\mathbf{X})$ is the discrepancy between the
target SDF values and the contact-derived mean evaluated at the constraint
positions. The coefficient vector $\boldsymbol{\alpha}$ determines how this
residual is spatially interpolated. At a query $\mathbf{x}$, the interpolated GP
residual is added back to $m(\mathbf{x})$~\cite{c9}. Consequently, the CSP guides weakly observed regions, while the GP residual modifies it to satisfy
the tactile constraints. It is not a hard shape constraint and does not force a
constant cross-section in the final mesh. When the mean function is disabled,
$m(\mathbf{x})=0$ and Eq.~\eqref{eq:gpis_nonzero_mean} reduces to the standard
zero-mean formulation.

We solve the linear system using Cholesky factorization~\cite{c12} without explicitly
forming $\mathbf{A}^{-1}$. If Cholesky factorization fails because of numerical
degeneracy, we solve the same system by least squares with
$\mathtt{rcond}=10^{-5}$. Only the posterior mean in
Eq.~\eqref{eq:gpis_nonzero_mean} is used for reconstruction; posterior variance
is not evaluated. Finally, we sample $f(\mathbf{x})$ on a $64^3$ volumetric grid
and extract its zero-level set using Marching Cubes~\cite{c13}. If multiple connected components are present,
we retain the component closest to the measured contacts. The complete pipeline
requires no iterative model training, pretrained shape model, text prompt, class
label, or object identifier.

%% file: 7_experiment_HW_setup.tex
\label{sec:Experiments}

\subsection{Hardware Setup}
\begin{figure}[!h]
    \centering
    \includegraphics[width=\columnwidth]{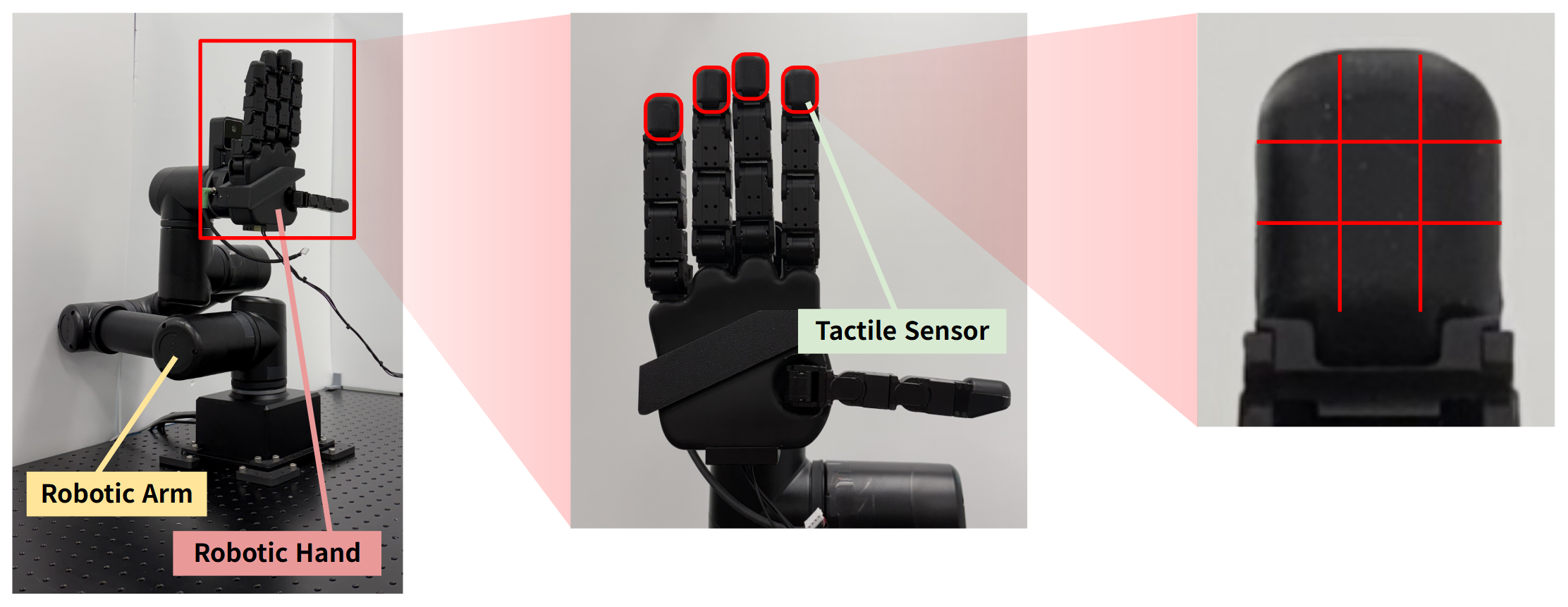}
    \caption{Hardware setup of the proposed system, showing the ROBOTIS OMY robotic arm, ROBOTIS HX5-D20-MRT robotic hand, and finger pad tactile sensing regions.}
    \label{fig:hardware_setup}
\end{figure}

Our experimental setup consists of a 6-DoF ROBOTIS OMY robotic arm equipped with a 20-DoF ROBOTIS HX5-D20-MRT robotic hand. A $3 \times 3$ tactile sensor array is mounted on the finger pad of each finger to capture local contact information during object interaction. Each sensor array consists of nine sensing elements, with each element providing an 8-bit raw sensor value ranging from 0 to 255. The experimental setup is shown in Fig.~\ref{fig:hardware_setup}.

%% file: 8_expriment_object.tex
\label{sec:Experiments}

\begin{figure}[!h]
    \centering
    \includegraphics[width=\columnwidth]{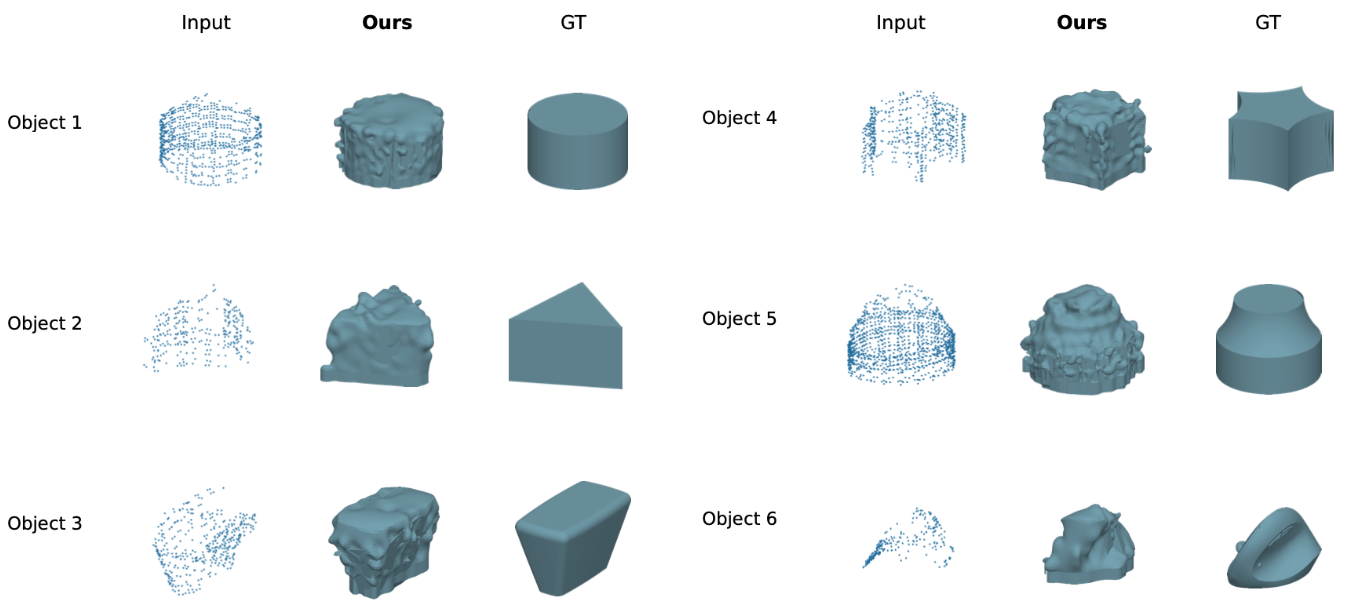}
    \caption{Input contacts, reconstructed meshes produced by
    the proposed method (Ours), and ground-truth (GT) reference
    meshes for Objects 1--6. Objects 1--3 are shown on the left
    and Objects 4--6 on the right. Within each object, the input,
    reconstruction, and GT are displayed from the same viewpoint
    and at the same scale.}
    \label{fig:mesh_reconstruction}
\end{figure}

\subsection{Experiment Objects}

We evaluated the proposed method using six objects.
Fig.~\ref{fig:mesh_reconstruction} shows their ground-truth (GT) reference meshes.
The objects had base widths of approximately 8--10 cm and heights
of up to 6 cm, chosen to accommodate the reachable workspace of
the robotic hand. Flat bottom surfaces were used to facilitate
stable mounting and consistent tactile exploration.

The physical test objects were fabricated using a 3D printer.
Each object incorporated a cylindrical mounting hole at the bottom,
measuring 20.24 mm in diameter and 25 mm in depth.
During tactile exploration, the objects were secured to a support
at a mounting height of approximately 20 cm.

%% file: 9_experiment_reconstruction.tex
\label{sec:Experiments}

\subsection{Mesh Reconstruction}

\begin{table}[t]
\centering
\vspace*{2mm}
\caption{Reconstruction performance on six objects.}
\label{tab:reconstruction_results}
\begin{tabular}{l|cccc}
\hline
\multicolumn{1}{c|}{Object} & CD$\downarrow$ & EMD $\downarrow$ & F-Score $\uparrow$ & Reconstruction time(s) \\
\hline
Object 1 & 1.589 & 7.993 & 96.98 & 61.67\\
Object 2 & 1.500 & 7.144 & 95.82 & 13.01\\
Object 3 & 1.602 & 6.817 & 95.71 & 34.10\\
Object 4 & 1.166 & 7.405 & 99.14 & 47.94\\
Object 5 & 1.204 & 7.038 & 97.86 & 127.60\\
Object 6 & 2.602 & 10.033 & 84.49 & 15.64\\
\hline
Average & 1.610 & 7.738 & 95.00 & 49.99\\
\hline
\end{tabular}
\end{table}
 
\begin{table}[t]
\centering
\caption{Quantitative comparison in the observed region.}
\label{tab:observed_region}
\setlength{\tabcolsep}{4pt}
\begin{tabularx}{\columnwidth}{>{\raggedright\arraybackslash}X|ccc}
\hline
\multicolumn{1}{c|}{Method} & CD $\downarrow$ & EMD $\downarrow$ & F-Score $\uparrow$ \\
\hline
Standard GPIS (Single-scale) & 3.448 & 11.060 & 75.580 \\
Multi-scale GPIS             & 1.244 & \textbf{6.491} & \underline{97.504} \\
Multi-scale GPIS + PCG       & 1.248 & \textbf{6.491} & \underline{97.504} \\
Multi-scale GPIS + CSP       & \textbf{1.229} & 6.818 & 97.497 \\
\hline
Ours: Multi-scale GPIS + CSP + PCG & \underline{1.236} & \underline{6.637} & \textbf{97.509} \\
\hline
\end{tabularx}
\end{table}

\begin{table}[t]
\centering
\vspace*{2mm}
\caption{Quantitative comparison in the unobserved region.}
\label{tab:unobserved_region}
\setlength{\tabcolsep}{4pt}
\begin{tabularx}{\columnwidth}{>{\raggedright\arraybackslash}X|ccc}
\hline
\multicolumn{1}{c|}{Method} & CD $\downarrow$ & EMD $\downarrow$ & F-Score $\uparrow$ \\
\hline
Standard GPIS (Single-scale)       & 4.076 & 13.019 & 71.253 \\
Multi-scale GPIS                   & 2.133 & 8.968  & 89.705 \\
Multi-scale GPIS + PCG             & 2.102 & \underline{8.731}  & 90.124 \\
Multi-scale GPIS + CSP             & \underline{2.003} & \textbf{8.573}  & \underline{92.148} \\
\hline
Ours: Multi-scale GPIS + CSP + PCG & \textbf{1.992} & 8.862  & \textbf{92.605} \\
\hline
\end{tabularx}
\end{table}

We evaluate reconstruction quality using Chamfer Distance
(CD) and Earth Mover's Distance (EMD) \cite{cCD_EMD}, and
F-Score \cite{cFScore}. CD and EMD are reported in mm, while F-Score is
reported as a percentage using a distance threshold of 5 mm.
For the reported reconstruction comparisons,
contact positions were projected onto the GT surface and
perturbed along the surface normals with Gaussian noise
of standard deviation 0.5 mm. These results therefore
characterize reconstruction under GT-assisted input conditions.
The input contacts and object-specific registration transforms
are fixed across methods.

Fig.~\ref{fig:mesh_reconstruction} shows the input contacts,
meshes reconstructed by the proposed method, and corresponding
GT meshes for Objects 1--6. The reconstructions capture the
overall object geometry, although local surface irregularities
and incomplete shape recovery remain visible.
Table~\ref{tab:reconstruction_results} summarizes the
object-wise quantitative results. The proposed method achieves
an average CD of 1.610 mm, EMD of 7.738 mm, and F-Score
of 95.00\%. Tactile exploration takes approximately 30 min
per object on average. The average recorded reconstruction
time is 49.99 s, including input loading, GP inference,
mesh extraction, boundary closure, and flat-bottom construction,
but excluding the separate filling stage and metric evaluation.

To examine the contributions of the proposed components,
we compare five configurations: Standard GPIS (Single-scale),
Multi-scale GPIS, Multi-scale GPIS + PCG,
Multi-scale GPIS + CSP, and the complete proposed method,
Multi-scale GPIS + CSP + PCG (Ours).
Standard GPIS uses a single RBF kernel with a normalized
length scale of 0.16, whereas Multi-scale GPIS combines
fine- and coarse-scale RBF kernels.
PCG and CSP are evaluated both individually and jointly.
Contact-reliability processing and the boundary-closure,
flat-bottom, and filling procedures are shared across
all five configurations.

We separately evaluate observed and unobserved regions.
A surface sample is classified as observed if it lies within
5 mm of an input contact; the remaining samples are classified
as unobserved. Generated pseudo-contacts are excluded from
this partition.
Tables~\ref{tab:observed_region} and
\ref{tab:unobserved_region} report the corresponding
region-wise results, averaged equally across the six objects.

Compared with Multi-scale GPIS, Ours reduces mean
unobserved-region CD from 2.133 to 1.992 mm and increases
F-Score from 89.705\% to 92.605\%.
Differences among the multi-scale configurations are smaller
in the observed region.
The improvements are metric-dependent:
Multi-scale GPIS + CSP achieves the lowest unobserved-region
EMD, whereas Ours achieves the lowest unobserved-region CD
and highest F-Score.

%% file: 11_conclusion.tex
 This paper presented a training-free and vision-free framework for 3D object reconstruction using tactile sensing alone. Contacts acquired from $3 \times 3$ pressure arrays on a multi-finger robotic hand mounted on a robotic arm are integrated using GPIS to reconstruct continuous object geometry without requiring visual observations, object class information, or learned generative shape priors, while the exploration policy adapts the number of scan layers to the measured object height. Experiments on six real objects demonstrated the feasibility of the proposed approach in a real-world robotic setting, while multi-finger sensing enables multiple contact measurements to be acquired within each exploratory action. Because no generative model is involved, reconstruction is completed in 49.99 s from the accumulated contacts. Future work includes replacing the fixed exploration sequence with an adaptive touch policy driven by surface uncertainty, and improving reconstruction quality from sparser contact sets so that fewer exploratory actions are required. By enabling both geometric perception and physical interaction on the same robotic platform, this work provides a practical foundation for reconstructing and manipulating unknown objects when visual sensing is unavailable. 

%% file: 12_Appendix.tex
\label{sec:apendix}

\subsection{ Visualization of PCG and CSP }

Fig.~\ref{fig:appendix_results} illustrates PCG and CSP construction for the object 3. Nearby contacts with compatible normals generate midpoint pseudo-contacts. Layer-wise side-contact hulls are then combined and expanded to form a shared cross-section, which defines the geometric GP mean.

\begin{table*}[t]
\centering
\caption{Overall Chamfer distance (mm, $\downarrow$) after Fill on six objects.}
\label{tab:cd_comparison}
\begin{tabular}{lccccccc}
\hline
Method & Object1 & Object2 & Object3 & Object4 & Object5 & Object6 & Mean \\
\hline
Single-scale GPIS
& 4.456 & 3.277 & 4.324 & 3.362 & 2.915 & 4.261 & 3.766 \\
Multi-scale GPIS
& 1.673 & 1.828 & 1.791 & 1.193 & 1.229 & \textbf{2.521} & 1.706 \\
Multi-scale GPIS+PCG
& 1.674 & 1.766 & 1.758 & 1.205 & 1.213 & 2.533 & 1.691 \\
Multi-scale GPIS+CSP
& \textbf{1.569} & 1.597 & 1.684 & \textbf{1.150} & 1.214 & 2.594 & 1.635 \\
Ours
& 1.590 & \textbf{1.552} & \textbf{1.680} & 1.166 & \textbf{1.204} & 2.602 & \textbf{1.632} \\
\hline
\end{tabular}
\end{table*}